\documentclass[10pt,twocolumn,letterpaper]{article}

\usepackage{cvpr}              % To produce the CAMERA-READY version
\usepackage[dvipsnames]{xcolor}

\definecolor{cvprblue}{rgb}{0.21,0.49,0.74}
\usepackage[pagebackref,breaklinks,colorlinks,citecolor=cvprblue]{hyperref}

\usepackage{multirow}
\usepackage{xspace}
\newcommand{\ours}{FT-CLIP\xspace}
\title{Improving Chinese Character Representation with Formation Tree}

\author{
Yang Hong\textsuperscript{\rm 1,2},
Yinfei Li\textsuperscript{\rm 2},
Xiaojun Qiao\textsuperscript{\rm 1},
Rui Li\textsuperscript{\rm 1},
Junsong Zhang\textsuperscript{\rm 2,3} \footnotemark[1]\\
\textsuperscript{\rm 1} NERCEL, Central China Normal University, China\\
\textsuperscript{\rm 2} Mind, Art and Computation Group, Department of Artificial Intelligence, Xiamen University, China\\
\textsuperscript{\rm 3} Key Laboratory of Digital Protection and Intelligent Processing of Intangible Cultural Heritage \\of Fujian and Taiwan (Xiamen University), Ministry of Culture and Tourism, China\\
{\tt\small\{hongyang, xj\_qiao\} @mails.ccnu.edu.cn, 
liyinfei@stu.xmu.edu.cn, 
leerui@ccnu.edu.cn} \\
{\tt\small zhangjs@xmu.edu.cn}
}

\begin{document}
\maketitle

{
  \renewcommand{\thefootnote}{\fnsymbol{footnote}}
  \footnotetext[1]{Corresponding author.}
}

\begin{abstract}
Learning effective representations for Chinese characters presents unique challenges, primarily due to the vast number of characters and their continuous growth, which requires models to handle an expanding category space. Additionally, the inherent sparsity of character usage complicates the generalization of learned representations. Prior research has explored radical-based sequences to overcome these issues, achieving progress in recognizing unseen characters. However, these approaches fail to fully exploit the inherent tree structure of such sequences. To address these limitations and leverage established data properties, we propose Formation Tree-CLIP (FT-CLIP). This model utilizes formation trees to represent characters and incorporates a dedicated tree encoder, significantly improving performance in both seen and unseen character recognition tasks. We further introduce masking for to both character images and tree nodes, enabling efficient and effective training. This approach accelerates training significantly (by a factor of 2 or more) while enhancing accuracy. Extensive experiments show that processing characters through formation trees aligns better with their inherent properties than direct sequential methods, significantly enhancing the generality and usability of the representations.
\end{abstract}
\section{Inroduction}
Chinese character representations, which convey the information in each Chinese character image, are crucial for various tasks, particularly character recognition. However, the long-tail effect hampers the performance of existing methods in handling rarely seen or even unseen characters, due to the diverse frequency of character usage in daily life~\cite{diao2023}. Consequently, there is a pressing need for a representation learning method that can effectively generalize across unseen characters to address these challenges.

Most existing Chinese character recognition methods~\cite{xiao2017, li2018, xu2019a} predict a fixed set of predetermined character categories, which limits their generality and usability since additional labeled data is required for characters outside the predefined set. To overcome these constraints, newer approaches incorporate radical-level details to handle unseen characters. RD-Net~\cite{wang2017}, for example, detects position-dependent radicals using multi-label learning. Similarly, RAN~\cite{zhang2020} and DenseRAN~\cite{wang2018} employ radical-based sequences and approach character recognition as a caption matching task. RZCR~\cite{diao2023} introduces a Knowledge Graph Reasoner to replace hard-matching strategies with reasoning-based approaches, achieving state-of-the-art performance. CCR-CLIP~\cite{yu2023} uses CLIP~\cite{radford2021} to align the embedding spaces for character images and radical-based sequences, significantly outperforming previous methods. However, these approaches typically treat radical-based sequences sequentially, neglecting their inherent tree structure, which could enhance recognition accuracy and efficiency.

\begin{figure}[!t]
    \centering
    \subfloat[Radical-based Sequence]{\includegraphics[width=0.2\textwidth]{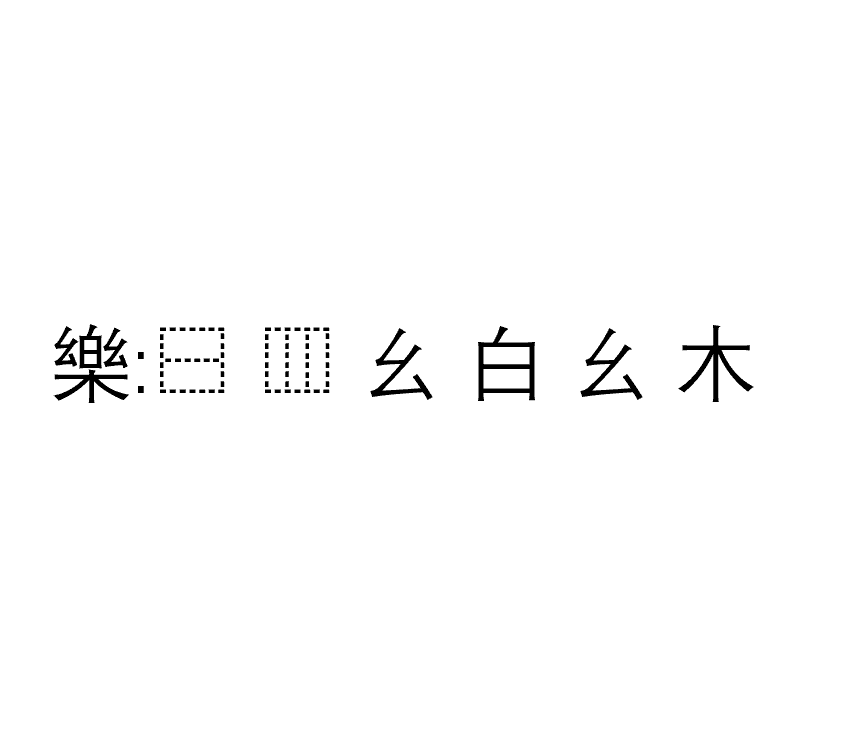}%
    \label{fig:seq_tree_ids}}
    \hfil
    \subfloat[Decomposition Tree]{\includegraphics[width=0.2\textwidth]{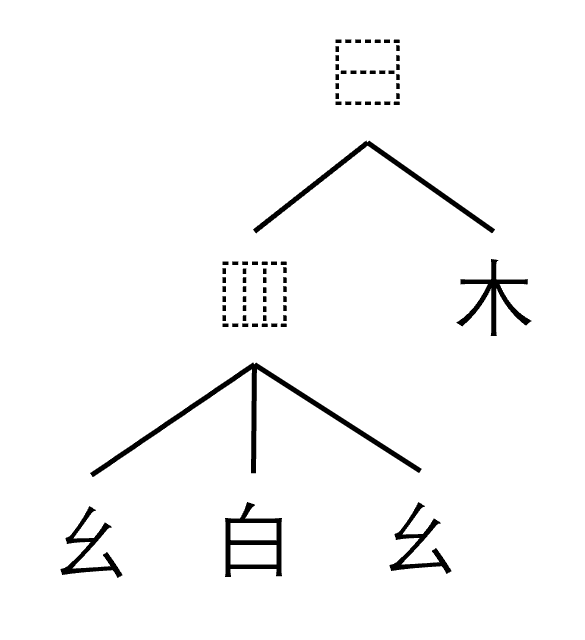}%
    \label{fig:seq_tree_dtree}}
    \hfil
    \subfloat[Sequential Approach]{\includegraphics[width=0.2\textwidth]{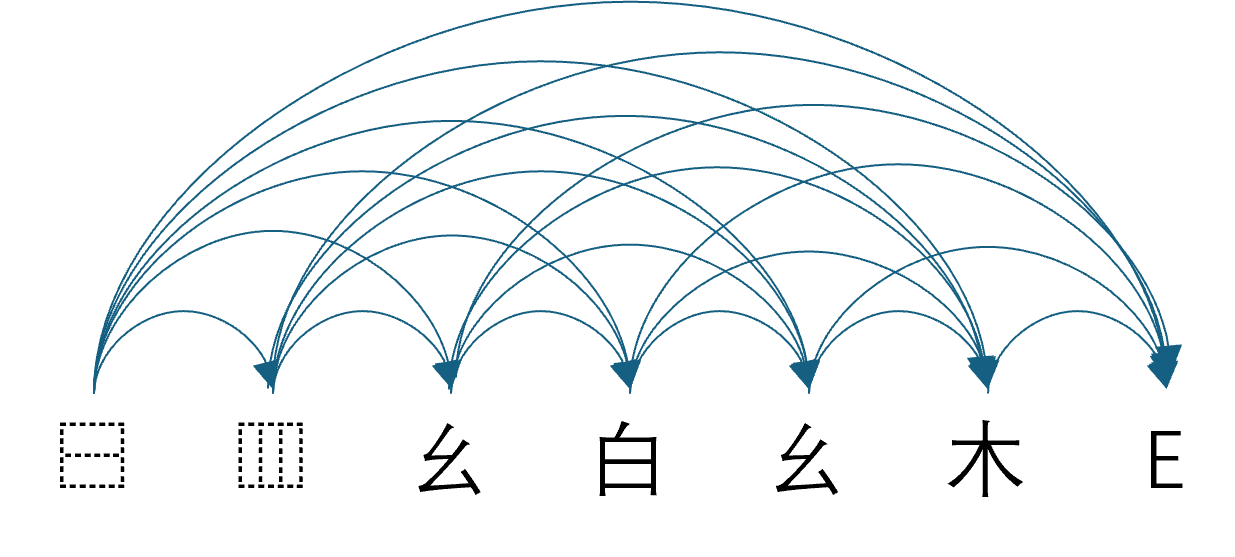}%
    \label{fig:seq_tree_seq}}
    \hfil
    \subfloat[Formation Tree]{\includegraphics[width=0.2\textwidth]{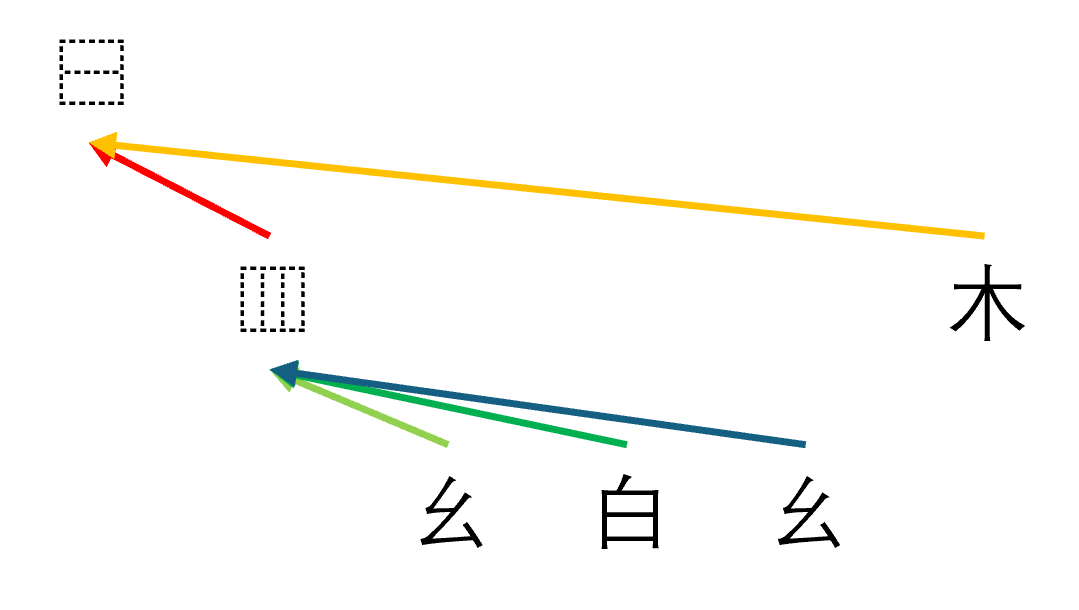}%
    \label{fig:seq_tree_ftree}}
    \caption{Illustration of radical-based sequences, which preserves the hierarchical knowledge in the form of the decomposition tree. Existing approach encode them in a sequential manner using look-ahead mask, while our approach transform the decomposition tree into a formation tree with fixed edge direction and formation type-related edge attributes.}
    \label{fig:seq_tree}
\end{figure}

To fully utilize the information conveyed by radical-based sequences, it's essential to consider the hierarchical knowledge of Chinese characters, which is effectively represented using decomposition trees as illustrated in Fig.~\ref{fig:seq_tree_dtree}. In this structure, radicals form the leaf nodes, while the parent nodes represent formation types that dictate the layout of these radicals relative to each other. Notably, these sequences are derived from decomposition trees through a depth-first traversal process. Current approaches like CCR-CLIP~\cite{yu2023} encode these sequences sequentially using a look-ahead mask, as shown in Fig.~\ref{fig:seq_tree_seq}, but overlook the underlying tree structure. HDE~\cite{cao2020} attempts to integrate hierarchical knowledge by focusing on node features and tree path attributes, yet it primarily addresses local features rather than the character as a whole. RTN~\cite{xue2023} employs tree-based positional encoding~\cite{shiv2019} within a Transformer framework~\cite{vaswani2017} to suggest connections between nodes, but it lacks control over feature aggregation. Furthermore, the restriction of these tree structures to binary trees, as determined by their positional relationships, curtails the diversity and usability of radical-based descriptive methods.

To more effectively learn character representations based on the properties of radical-based sequences, we convert the inherent decomposition trees into formation trees as depicted in Fig.~\ref{fig:seq_tree_ftree}. The decomposition tree disassembles a character into its constituent radicals, whereas learning whole character features from these radicals is the reverse process. To facilitate this, we first ensure the edges in the formation tree point from each child node to its parent node, aligning with the direction of feature aggregation from the radicals to the whole character. Furthermore, each child node under a specific formation type possesses unique locational attributes, such as the left part in a left-right formation, which we refer to as the azimuth. The type of each edge corresponds to the azimuth of its child node and is visually differentiated by various colors. Importantly, we have developed a custom tree encoder that learns character representations by effectively leveraging the formation tree’s structural characteristics.

Most studies utilize graph neural networks~\cite{velickovic2018, hamilton2017, kipf2017} to learn representations of non-Euclidean data; however, these networks often struggle with challenges such as over-smoothing and difficulty in modeling long-distance relationships~\cite{ju2024}. Graph Transformers~\cite{dwivedi2021, ying2021, zhang2022a} offer a compelling alternative, demonstrating superior performance across a variety of graph representation learning tasks. In light of this, we have adapted the standard Transformer architecture~\cite{vaswani2017} to better suit the unique characteristics of formation trees, incorporating two simple yet effective encoding methods. To align the formation tree representation more closely with its corresponding image representation, we use the contrastive language-image pretraining (CLIP) framework to simultaneously train our tree encoder and a Vision Transformer (ViT)~\cite{dosovitskiy2020}. Both encoders, being based on the Transformer model, benefit from the incorporation of masking techniques, which are tailored to the specifics of character images and formation trees, thus speeding up training and improving accuracy. The essence of our CLIP-based approach is the effective utilization of the Formation Tree of Chinese characters, which we refer to as \ours in our approach.

In summary, main contributions of the proposed \ours are summarized as follows:
\begin{enumerate}
\item We propose a novel formation tree to represent Chinese characters with radical-level details. It conceptualizes the formation of Chinese characters as a hierarchical aggregation of radical-based features, progressing from the leaf nodes up to the root node. This is achieved by constraining the direction of feature aggregation and defining the spatial relationships of related components using edge relationships. Our representation method is concise and effective, and it conforms to the characteristics of Chinese characters.
\item We propose a novel tree encoder for the formation tree with two simple yet effective encoding methods. One method incorporates edge features into the process of feature aggregation by modifying the attention operation, and the other defines a specialized spatial location encoding based on the spatial location relationship of leaf nodes corresponding to their parent nodes. Extensive experiments have proved the effectiveness of both methods.
\item We investigated the utilize of masks in the learning of Chinese character representations, both in images and tree nodes, where masks not only improve training efficiency but also recognition accuracy.
\end{enumerate}

\section{Related work}

\subsection{Chinese Character Recognition}
Deep learning based methods have achieved the state-of-the-art performance for Chinese character recognition by learning discriminative representations directly from raw data.
Traditional approaches have given way to deep learning-based approaches, with convolutional neural networks (ConvNets)~\cite{lecun1998} being particularly noteworthy. 

\paragraph{Traditional Approaches}
Before the advent of deep learning, traditional methods for Chinese character representation can be categorized into three groups: character-based approaches, radical-based approaches, and stroke-based approaches.
Character-based approaches utilize hand-crafted features such as Gabor features~\cite{su2003}, directional features~\cite{jin2001} and vector features~\cite{chang2008}. 
However, these approaches do not effectively incorporate Chinese character domain knowledge and their performance is limited by the low-capacity features~\cite{chen2022}. 
Radical-based approaches utilize the hierarchical component decomposition process of Chinese characters, and the recursive radical matching scheme~\cite{wang2001,wang1996} is widely used. 
However, these methods are time-consuming and may face limitations in performance when applied to datasets with large-scale character sets.
Stroke-based approaches are based on strokes, the smallest units of Chinese characters, and thus are widely applied in traditional methods. 
Stroke-based Attributed Relational Graph (ARG) is one of the most successful methods to represent Chinese characters. 
In ARG, a node corresponds to a stroke in the Chinese character, and edges represent the inter-stroke relation defined by different authors. 
ARG has been applied in character recognition~\cite{liu2001}, but it is ineffective and inefficient when dealing with complex characters. 
In general, traditional methods for character recognition can only achieve good performance on a small range of characters and are less effective when dealing with large character sets, let alone unseen characters.

\paragraph{Deep Learning-based Approaches}
The Multi-column deep neural network is the first successful attempt at achieving convincing performance for character recognition~\cite{ciresan2015}. 
Since then, ConvNet-based methods have been proposed and achieved even better performance through advanced network architecture~\cite{xu2019a,xiao2017,li2018} and improved training strategies~\cite{wu2014,chen2015}. 
However, these approaches are limited in their generality and usability for unseen characters as they are mainly trained to predict a fixed set of predetermined character categories. 

Efforts have been made towards zero-shot recognition of unseen characters, which aims to overcome the limitations of discrete labels and recognize characters by analyzing their radicals and structures. 
RD-Net~\cite{wang2017} detects position-dependent radicals using a deep residual network with multi-labeled learning. 
HDE~\cite{cao2020} proposes a novel zero-shot hierarchical decomposition embedding method that encodes the tree layout into a semantic vector and classifies novel characters based on the compatibility between the handwriting embedding and the semantic embedding.
Some methods generate caption-like description sequences and search for the most similar descriptions in a predefined dictionary to recognize unseen characters.
DenseRAN~\cite{wang2018} and RAN~\cite{zhang2020} employ ConvNets as encoders to learn Chinese character image representations, while RNNs are employed as decoders to predict radical-based sequences symbol by symbol.
CUE~\cite{luo2023} is proposed to measure the importance of radicals in recognizing Chinese characters, with a novel Chinese character uncertainty elimination (CUE) framework to alleviate the radical sequence mismatch problem.
The stroke-based method~\cite{chen2021} is proposed with an encoder with ConvNets and a decoder with Transformer, achieving state-of-the-art performance in zero-shot Chinese character recognition tasks by solving radical zero-shot cases with stroke-based sequences.
Inspired by these methods, STAR~\cite{zeng2023} combined stroke- and radical-level decompositions to achieve better performance.
These sequence-based methods employed a hard-matching strategy to match the generated sequences and dictionary, which limits their applications in practice.
RZCR~\cite{diao2023} proposes Knowledge Graph Reasoner to replace hard-matching strategies with a reasoning-based strategy, and the state-of-the-art performance is achieved.
CCR-CLIP~\cite{yu2023} utilize CLIP to aligns the two embedding spaces learned for character images and radical-based sequences respectively, and outperforms previous methods with a decent margin.
However, they handle radical-based sequences in a sequential manner, ignoring their inherent tree structure.
Existing approaches treat radicals and structures as auxiliary annotations rather than effectively integrating the relationships between them and treat them as a whole, which limits their performance when dealing with complex characters.

\subsection{Graph Transformers}
Graph Neural Networks~\cite{velickovic2018,hamilton2017,kipf2017}, while excelling in processing non-Euclidean data through the message-passing paradigm, encounter inherent challenges such as over-smoothing and long-distance modeling~\cite{ju2024}. To address these issues, one notable strategy has been to leverage the Transformer architecture~\cite{vaswani2017} for graph representation learning. The core mechanism of the vanilla Transformer, which utilizes an attention operation to facilitate information exchange based on the similarity between source and target elements, is quite similar to the message-passing process observed on a fully-connected graph. However, applying this framework directly to arbitrary graphs without considering their structural information may lead to poor performance, particularly when graph topology is important. Moreover, positional encoding in graphs is not a trivial problem because the order or coordinates of graph nodes are underdefined. Given these two challenges, Transformer-based methods for graph representation learning can be broadly divided into two categories: one considering graph structure through \textbf{Modified Attention Operation}, and the other embeds the topological information of the graph for \textbf{Enhanced Node Features}.

\paragraph{Modified Attention Operation}
Graph Transformer~\cite{dwivedi2021} restrict node features to attend solely to neighboring nodes, implicitly encoding edge scores through pairwise attention. HGT~\cite{hu2020} disentangles the attention of different node types and edge types by adopting additional attention heads. HetGT~\cite{yao2020} defines four types of subgraphs to capture global, undirected, forward and backward information respectively. Path features between nodes are consistently integrated as inductive bias added to the original score function. Graph Transformer~\cite{cai2020} employs GRU~\cite{chung2014} to encode forward and backward features. Graphormer~\cite{ying2021} introduces structural features as an attention bias by utilizing both the length of paths and path embeddings. GRIT~\cite{ma2023} leverages relative random walk probabilities as an inductive bias to encode information about the relative paths between nodes. Graphormer-GD~\cite{zhang2022a} also incorporates relative distance as a bias and provides a rigorous proof of the significance of this bias in determining the biconnectivity of a graph.

\paragraph{Enhanced Node Features}
Graph Transformer~\cite{dwivedi2021} incorporates the eigenvectors associated with the k smallest non-trivial eigenvalues to create intermediate embeddings, which are then mapped into a d-dimensional space to derive positional encodings. GTSA~\cite{kreuzer2021} on the other hand, employs sequence modeling techniques to generate positional encodings, offering a different approach. Given that the Laplacian eigenvectors may yield complex values for directed graphs, EGT~\cite{hussain2022} using the Singular Value Decomposition (SVD) of the adjacency matrix as an alternative. Furthermore, the Digraph Transformer~\cite{geisler2023} innovatively applies the Magnetic Laplacian to process directed graphs. For tree-structure data, \cite{shiv2019} extended the sequential positional encoding to tree-based positional encoding in Transformer, a method also utilized in radical-based sequence encoding by RTN\cite{xue2023}. TPTrans\cite{peng2021} innovates by encoding both the pairwise path among tokens in source code and the path from each token to the tree root in the syntax tree, enhancing code representation learning with a deep understanding of structural relationships. In contrast to position-analogy approaches, structure-aware approaches do not seek to rigorously emulate sequential positional encoding through mathematical means. Graphormer~\cite{ying2021} proposes to leverage node centrality as an additional feature to address the importance of each node. Graph-BERT~\cite{zhang2020b} employs Weisfeiler–Lehman algorithm to label each node.

It is worth noting that the relationships between nodes in trees and graphs have clear differences. For example, in trees, the edges connect parent nodes and child nodes, and there is a clear hierarchical relationship between them. Moreover, unlike the ambiguous node position features in graphs, the child nodes of the same parent node have a clear order relationship. In this paper, to better exploit the the tree structure, we modified Transformer in both attention operation and node features based on the characteristics of our formation tree.

\begin{figure}[h]
  
  \centering
  \includegraphics[width=\linewidth]{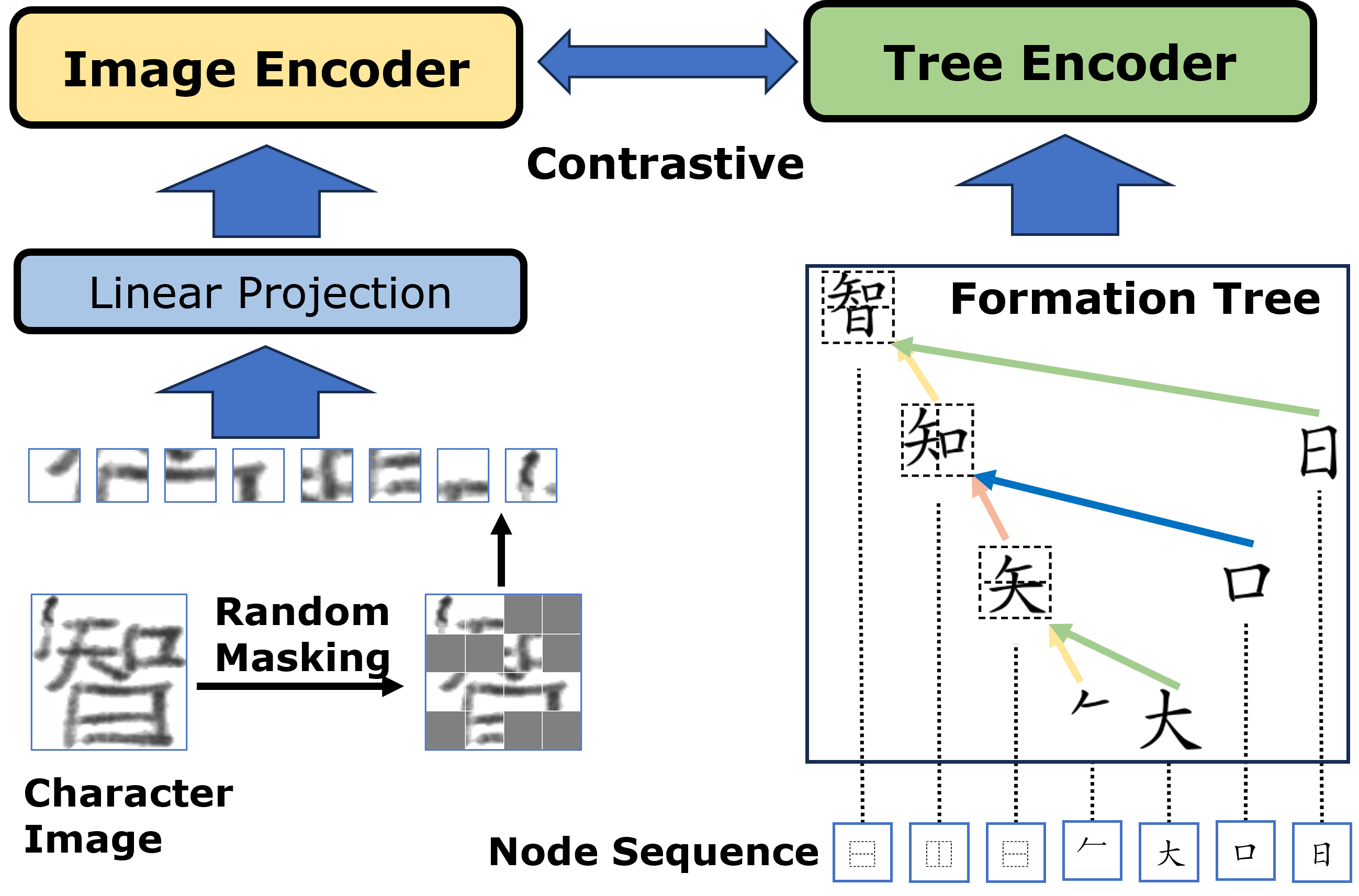}
  \caption{\ours jointly trains Image Encoder and Tree Encoder to predict the correct pairings of a batch of training examples.
  ViT is employed as image encoder with random masking, a novel Formation-Tree Transformer is proposed as tree encoder.}
  \label{fig:ftclip}
\end{figure}

\section{Proposed methodology}
As illustrated in Fig.~\ref{fig:ftclip}, the \ours model comprises two distinct encoders: one dedicated to character images and another to formation trees. Specifically, we employ a modified Vision Transformer (ViT) as our image encoder, complemented by a newly proposed Formation-Tree Transformer as our tree encoder. To effectively harness the inherent tree structure of radical-based sequences, these sequences are first transformed into formation trees prior to being fed into the tree encoder. Additionally, we utilize random masking in the image encoder during training, a technique that significantly speeds up the training process and concurrently enhances the accuracy of the model.

\begin{figure*}[!t]
    \centering
    \includegraphics[width=\linewidth]{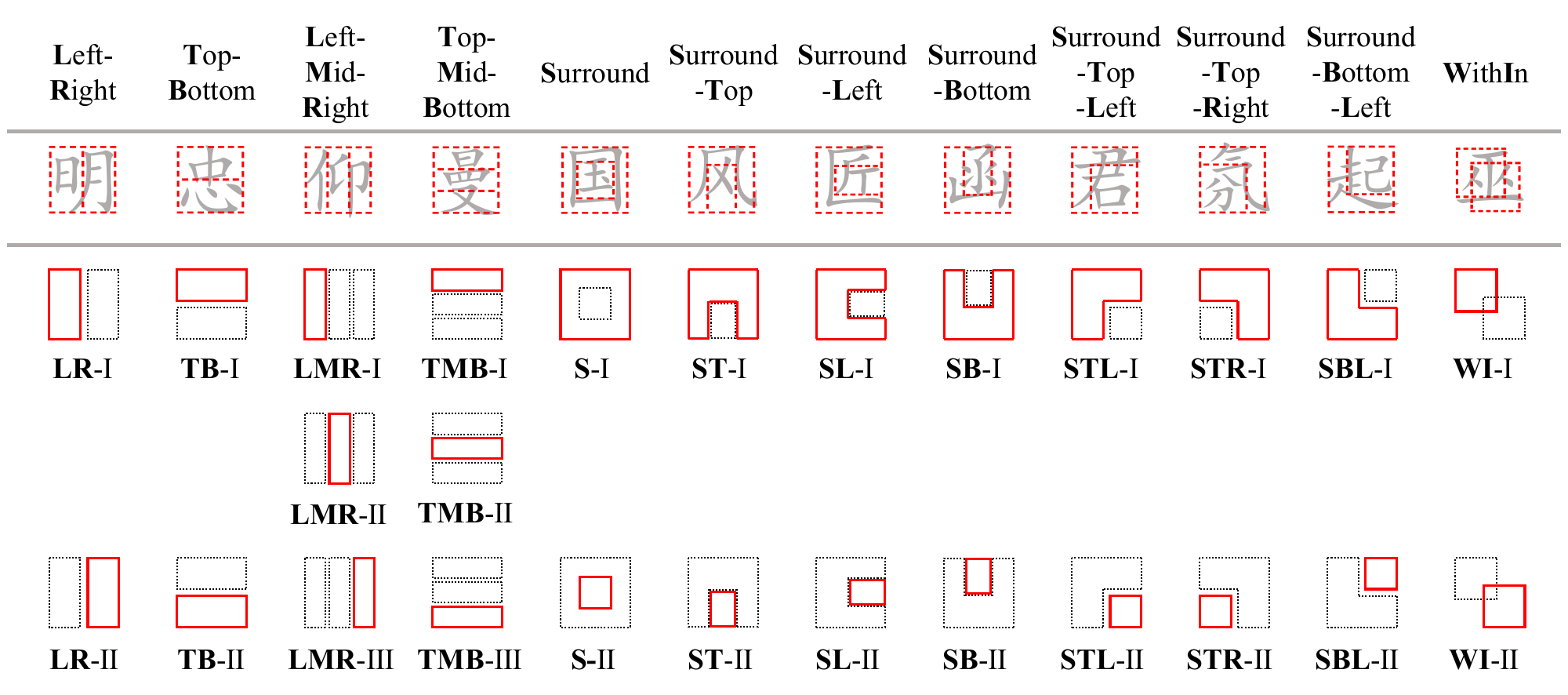}
    \caption{Twelve formation types used in \ours are defined in the top line, while the corresponding examples are illustrated below them.
        The bottom illustrates the azimuths as the red part of each corresponding formation type.
        % With two formation patterns describing spatial relation between three components, and the others describe the relationship between two, there are a total of 26 azimuths in our Chinese character formation.
        Azimuth names are defined based on the abbreviations of their corresponding formation type and their indexes in corresponding formations.}
    \label{fig:formation}
\end{figure*}

\subsection{Formation Tree}
Chinese character formation, also known as Chinese character structure~\cite{bao2015}, details how components are arranged relative to each other. 
Two prominent standards for character formation exist in prior research: Ideographic Description Sequences(IDS)~\cite{zhang2020,yu2023,cao2020} standard and QXK~\cite{devlin2019} standard. The IDS standard defines 12 formation patterns, while the QXK standard expands upon this by introducing 17 patterns encompassing more intricate structures. To ensure a consistent comparison with existing zero-shot Chinese character recognition studies\cite{zhang2020,yu2023,cao2020}, which predominantly utilize the IDS standard, we adopt this framework in our methodology.
As depicted in Fig.~\ref{fig:formation}, our approach incorporates all twelve IDS formation types. Notably, patterns like "Left-Middle-Right" and "Top-Middle-Bottom" capture the spatial relationships between three components, while the remaining types describe two-component relationships. This framework, further detailed in the bottom of Fig.~\ref{fig:formation}, identifies a total of 26 distinct azimuths. 
It's worth noting that some prior research~\cite{cao2020,xue2023} leverages binary tree structures for representing character formation, which necessitates the exclusion of complex types like "Left-Mid-Right" and "Top-Mid-Bottom." In contrast, our approach maintains compatibility with a wider range of formation types, fostering its generalizability across diverse character structures.

As illustrated in the lower right quadrant of Fig.~\ref{fig:ftclip}, the nodes within the formation tree are organized via a deep-first traversal, with each edge type determined by the azimuth of the corresponding child node and distinctively marked by various colors. All edges in the tree are directed, pointing from child nodes towards their parent nodes. This directional setup represents the sequence in which the character components are assembled to form the complete character, highlighting the hierarchical nature of Chinese character construction.

\subsection{Tree Encoder}
\begin{figure}[h]
  \centering
  \includegraphics[width=\linewidth]{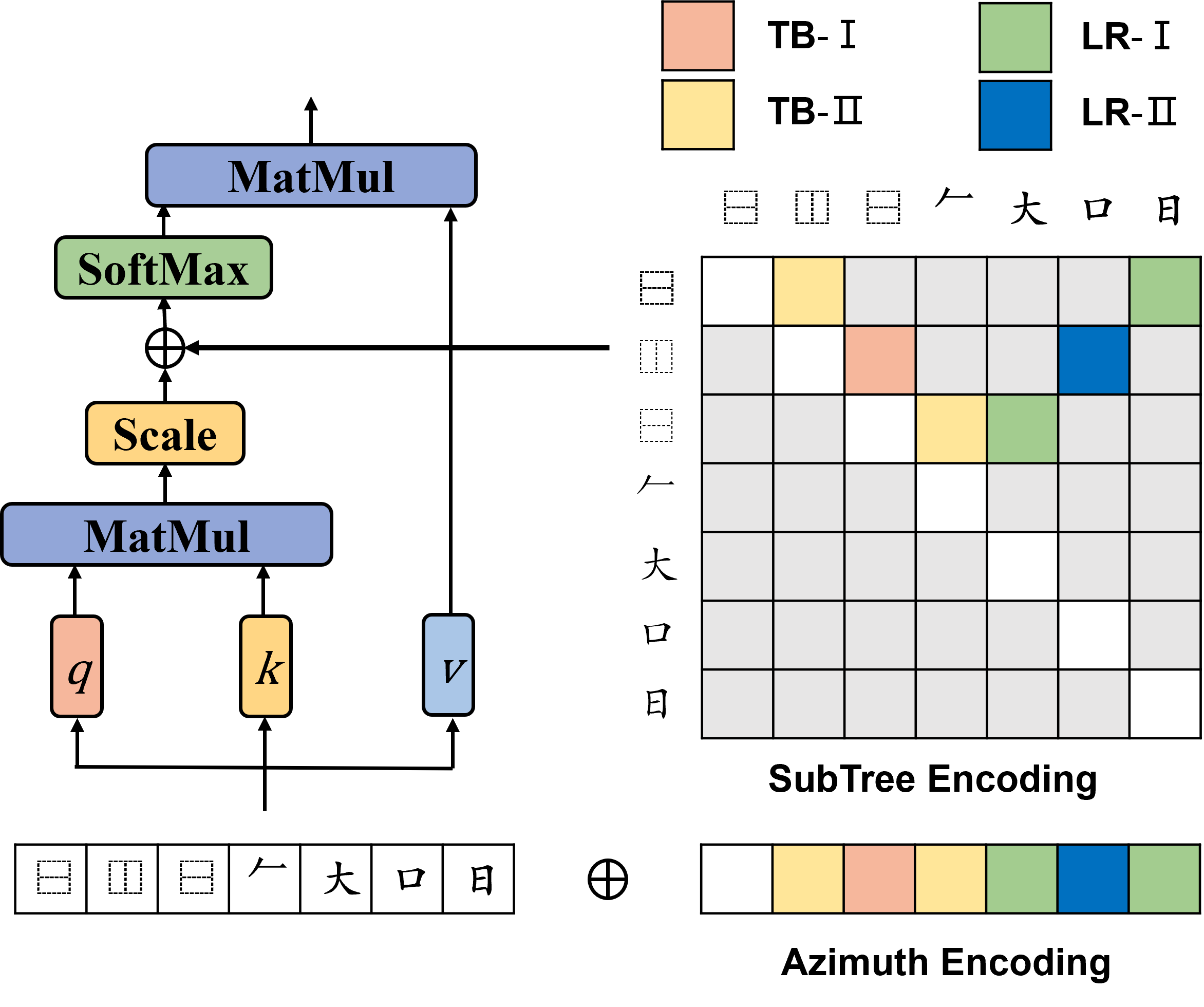}
  \caption{An illustration of the proposed SubTree Encoding and Azimuth Encoding in our tree encoder. 
  SubTree Encoding limits the self-attention to a node and its direct children, ignoring all other nodes (marked with grey).
  The type of edge between a node and its children is defined by the azimuth of the child node.
  Azimuth Encoding utilizes the azimuth of each node, as an additional feature of the node embedding. 
  % Corresponding azimuths are marked with different colors and are consistent with the edges in the previous figure.
  }
  \label{fig:tree_encoder}
\end{figure}

To better learn the representation of formation trees, two simple yet efficient encoding methods is proposed to fully capture the intrinsic properties of trees as shown in Fig.~\ref{fig:tree_encoder}.
We will recap the architecture of the Transformer first, and then detail the two encoding methods.

\paragraph{Transformer}
The Transformer architecture is consist of multiple Transformer layers~\cite{vaswani2017}.
Each layer has two modules: a self-attention module and a multi-layer perceptrons module.
Let $ H=\left[h_1^T,\ldots,h_n^T\right]\in\mathbb{R}^{n\times d} $ denote the input of self-attention modele where $d$ is the dimension and $h_i\in\mathbb{R}^{1\times d}$ is the hidden representation at position $i$.
We use three matrices ${W_Q\in\ \mathbb{R}}^{d\times d_k}$, ${W_K\in\ \mathbb{R}}^{d\times d_k}$, ${W_V\in\ \mathbb{R}}^{d\times d_V}$ to project input $H$ to the corresponding representation $Q$, $K$, $V$.
The self-attention is then calculated as:

\begin{equation}
    Q=HW_Q, K=HW_K, V=HW_V
\end{equation}

\begin{equation}
    A=\frac{QK^T}{\sqrt{d_k}}, Attn(H)=softmax(A)V
\end{equation}

where $A$ is a matrix capturing the similarity between queries and keys.

\paragraph{SubTree Encoding}
Instead of take advantage of the global receptive field in Transformer, SubTree Encoding limits the perceptual field to to nodes and their children, i.e., the smallest subtree of each parent node.
This restriction guides the aggregation of features from leaf nodes to the root node, which is consistent with the radical-based formation of Chinese characters.
Moreover, the formation type of the subtree is determined by the value of the parent node, and the type of each edge corresponds to the azimuths of the formation.
% Moreover, the subtrees correspond to various formation types, which are represented by the types of the parent nodes.
% In each formation pattern, each child node has distinct azimuth, which is represented by the type of edge between the child node and its parent.
SubTree Encoding incorporates edge features via a bias term to the attention module.
When calculate the similarity between a node and its children, we assign a learnable scalar as a bias term in the self-attention function to incorporate the features of the edges.
Concretely, we denote $A_{p,c}$ as the $\left(p,c\right)$-element of the Scaled Dot-Product matrix $A$ and $\left(p,c\right)$ is a parent-child node pair:
\begin{equation}
    A_{\left(p,c\right)}=\frac{\left(h_p\ W_Q\right)\left(h_c\ W_K\right)^T}{\sqrt d}+b_{azimuth\left(c\right)}
\end{equation}

Where $b_{azimuth\left(c\right)}$ is a learnable embedding vector specified by the type of edge between $\left(p,c\right)$, which is define by the azimuth of child node $c$ as $azimuth\left(c\right)$.
As for leaf nodes, they calculate self-attention on themselves.

\paragraph{Azimuth Encoding}
Positional encoding plays an important role in processing sequences, but not in graphs as there is no clear order between the nodes.
However, as a special type of graph, our formation tree has an explicit order among the child nodes of each subtree, corresponding to the azimuths under different formation types.
From the perspective of formation tree, using SubTree Encoding is sufficient to obtain unambiguous representations. 
However, we found that utilizing our Azimuth Encoding is beneficial in improving performance through experiments.
A possible explanation for this may be that Azimuth Encoding provides a basis for the deformation of radicals before the formation process. 
For example, radicals will become narrower under the left-right structure and flatter under the top-bottom structure.
% To enable the tree encoder to notice this phenomenon in the Chinese character images, we propose Azimuth Encoding to enable our encoder to recognize the difference of the same radical under different azimuths.
In our tree encoder, Azimuth Encoding assigns each node a real-valued embedding vector according to its azimuth types.
The Azimuth Encoding is applied to each node, and we add it to the node features before feed into the tree encoder.
Specifically, we use Azimuth Encoding to provide additional azimuth information as follows:

\begin{equation}
    h_c^{\left(0\right)}=x_c+z_{azimuth\left(c\right)}
\end{equation}

where $z \in \mathbb{R}^{1\times d}$ is a learnable embedding vector specific to the azimuth type of child node $c$.
% By using the azimuth encoding, the tree encoder can capture the difference of the same radical under different structures.
% Since the root node has no parent, Azimuth Encoding does not operate on the root node.

\paragraph{Special Node}
Most of the existing approach utilize a special node to learn the global representation.
In the BERT model~\cite{devlin2019}, it is [CLS], which is a special token attach at the beginning of the input sequences.
As for Graphormer~\cite{ying2021}, they add a special node [VNode] to the graph, and make connection between it and each node individually.
However, different from existing approaches, tree inherently has a root node, which, intuitively, is suitable for global representation learning.
Moreover, the root node in each formation tree corresponds to the feature of the whole character, since the subtree of the root node describes the final step of the radical-based Chinese character formation.
Therefore, instead of adding a special node, we use the root node to learn the global representation.
Ablation studies indicates that there is no need to add a special node to learn the global representation.

\subsection{Image Encoder}
We use ViT~\cite{dosovitskiy2020} as our image encoder instead of ResNet~\cite{he2016} as in CCR-CLIP~\cite{yu2023}.
ResNet is a widely adopted feature extractor and plays a crucial role in image representation learning.
However, using ViT as an image encoder outperforms significantly than using ResNet, as demonstrate in the experiments from original CLIP search~\cite{radford2021}.
Meanwhile, another advantage of using Transformer-based image encoder is that random masking can be employed to accelerate training efficiency while improving performance.
We split the Chinese character image into $8 \times 8$ patches, and then project each patch into 512-dimensional tokens, before feeding them into the Transformer layers.

\subsection{Masked Encoders}
Masking plays a crucial role in ensuring the proper functioning of the self-attention mechanism within Transformer models.
It was proposed by BERT~\cite{devlin2019} for learning deep contextual representation, then introduced to the field of computer vision by MAE~\cite{he2022}.
We employ masking in both our image encoder and tree encoder.

\paragraph{Image Masking}
In a nutshell, we masked out 50\% of image patches during training, which introduce a trade-off between "how we look closely at a character image" and "how many character images we can process in a batch."
Experimental results demonstrate that our method has an advantage in both aspects.
It worth noting that, \ours achieves the best performance with the mask ratio of 50\%, which is different from the optimal mask ratio of 75\% proposed by MAE.
The reason for this may be that inter-component correlations within the Chinese character images are more subtle than in a natural image.
A large mask ratio will make it difficult to identify Chinese characters.

\paragraph{Tree Masking}
During the inference process, we address the challenge of unseen leaf nodes in the formation tree by masking them out, ensuring they do not interfere with model predictions. To manage unseen radicals effectively, we explored three strategies: 1) Utilizing untrained random initial values as embeddings for these radicals; 2) Introducing a learnable [unk] node to substitute for unseen radicals, necessitating random replacement of a percentage of leaf nodes during training; 3) Completely ignoring unseen leaf nodes by applying masking. Attempts to enhance radical zero-shot recognition by replacing leaf nodes with [unk] nodes or masking leaf nodes during training did not yield the expected improvements. 
However, we observed that masking unseen radicals to exclude their influence on tree structure during inference significantly aids in leveraging known radical features to accurately match image-tree pairs. 
This approach simplifies the process and enhances the model’s ability to generalize to unseen characters.

\section{Experiments}
In this section, we will introduce all experiments conducted to verify the effectiveness of \ours on both seen and unseen Chinese characters.
We will introduce the used datasets at first.
Then, we will compare the results with related methods on the mentioned datasets.

\subsection{Dataset and Evaluation Metrics}
\paragraph{Dataset}
We conducted experiments on three widely used Chinese character recognition datasets, including CASIA-HWDB~\cite{liu2013}, ICDAR2013~\cite{yin2013} and CTW~\cite{yuan2019}.
While performing handwritten recognition tasks, samples from the CASIA-HWDB dataset are used as the training set, and those from the ICDAR2013 dataset are used as the test set.

\paragraph{Evaluation Metrics}
Following the previous Chinese character recognition works~\cite{zhang2020, cao2020, chen2021, yu2023}, we select the character accuracy as the evaluation metric.
During the inference phase, we initiate the process by calculating the tree representations for each character. Following this, we compute the representation of the image, which is then compared against these pre-computed tree representations. This comparison helps us identify the tree representation that most closely matches the image. Based on this closest match, we then determine the category of the image. 

\paragraph{Implementation Details}
Our code is written in Python and the network model is implemented using PyTorch, and all experiments are conducted on single NVIDIA RTX 4090 GPU with 24GB memory.
The image size is set to $32 \times 32$ for fair comparison with previous methods~\cite{chen2021, yu2023}. 
In both image encoder and tree encoder, the number of Transformer encoder layers is empirically set to 8, with the hidden size of 512.
We also utilize mixed precision to accelerate training while reducing video memory consumption, allowing us to set the batch size to 2048.

\begin{table*}[!t]
    \renewcommand{\arraystretch}{1.2}
    \centering
    \small
    \resizebox{\textwidth}{!}{
    \begin{tabular}{|l|llllll|lllll|}
    \hline
    &\multicolumn{6}{c|}{Character Zero-shot}                               & \multicolumn{5}{|c|}{Radical Zero-shot}            \\ \cline{2-12} 
    \multicolumn{1}{|c|}{}                                              &Train Set                   & 500             & 1000           & 1500           & 2000              & 2755              & 50              & 40             & 30             & 20                & 10                \\ \hline
    \multicolumn{1}{|c|}{\multirow{5}{*}{\shortstack{Hand-\\written}}}  &DenseRAN~\cite{wang2018}    & 1.70\%          & 8.44\%         & 14.71\%        & 19.51\%           & 30.68\%           & 0.21\%          & 0.29\%         & 0.25\%         & 0.42\%            & 0.69\%            \\
    \multicolumn{1}{|c|}{}                                              & HDE~\cite{cao2020}          & 4.90\%          & 12.77\%        & 19.25\%        & 25.13\%           & 33.49\%           & 3.26\%          & 4.29\%         & 6.33\%         & 7.64\%            & 9.33\%            \\
    \multicolumn{1}{|c|}{}                                              & SDN~\cite{chen2021}         & 5.60\%          & 13.85\%        & 22.88\%        & 25.73\%           & 37.91\%           & 5.28\%          & 6.87\%         & 9.02\%         & 14.67\%           & 15.83\%           \\
    \multicolumn{1}{|c|}{}                                              & STAR~\cite{zeng2023}        & 7.54\%          & 19.47\%        & 27.79\%        & 35.53\%           & 43.86\%           & 6.95\%          & 12.28\%        & 14.74\%        & 18.37\%           & 23.23\%           \\
    \multicolumn{1}{|c|}{}                                              & CUE~\cite{luo2023}          & 7.43\%          & 15.75\%        & 24.01\%        & 27.04\%           & 40.55\%           & $-$             & $-$            & $-$            & $-$               & $-$               \\
    \multicolumn{1}{|c|}{}                                              & CCR-CLIP~\cite{yu2023}      & 21.79\%         & 42.99\%        & 55.86\%        & 62.99\%           & 72.98\%           & 11.15\%         & 13.85\%        & 16.01\%        & 16.76\%           & 15.96\%           \\  \cline{2-12}
    \multicolumn{1}{|c|}{}                                              & \ours                       &\textbf{26.17\%} &\textbf{50.68\%}&\textbf{62.25\%}&\textbf{70.39\%}   &\textbf{78.96\%}   &\textbf{12.34\%} &\textbf{16.66\%}&\textbf{21.51\%}&\textbf{24.65\%}  &\textbf{27.8\%}     \\  \hline
    \multirow{5}{*}{Scene}                              & Train Set           & 500             & 1000           & 1500           & 2000           & 3150           & 50              & 40             & 30             & 20               & 10     \\ \cline{2-12}
    & DenseRAN~\cite{wang2018}  & 0.15\%          & 0.54\%         & 1.60\%         & 1.95\%         & 5.39\%         & 0\%             & 0\%            & 0\%            & 0\%              & 0.04\%  \\
    & HDE~\cite{cao2020}      & 0.82\%          & 2.11\%         & 3.11\%         & 6.96\%         & 7.75\%         & 0.18\%          & 0.27\%         & 0.61\%         & 0.63\%           & 0.90\%  \\
    & SDN~\cite{chen2021}      & 1.54\%          & 2.54\%         & 4.32\%         & 6.82\%         & 8.61\%         & 0.66\%          & 0.75\%         & 0.81\%         & 0.94\%           & 2.25\%  \\
    & STAR~\cite{zeng2023}      & 1.19\%          & 3.77\%         & 8.04\%         & 11.00\%         & 11.27\%         & 2.16\%          & 2.33\%         & 2.76\%         & 4.81\%           & 5.35\%  \\
    & CCR-CLIP~\cite{yu2023}      & \textbf{3.55\%}         & 7.70\%         & 9.48\%         &  17.15\%         & 24.91\%         & 0.95\%          & 1.77\%         & 2.36\%         & 2.59\%           & 4.21\%  \\ \cline{2-12}
    & \ours                & 1.8\%  & \textbf{9.17\%}&\textbf{15.69\%}&\textbf{24.36\%}&\textbf{32.39\%}& \textbf{2.98\%} & \textbf{3.68\%}& \textbf{4.42\%}& \textbf{5.82\%}  & \textbf{8.13\%} \\ \hline
    \end{tabular}
    }
    \caption{Results of character zero-shot (left column) and radical zero-shot (right column) tasks on Handritten characters (first row) and Scene characters (last row). 
    Train Set indicates the total number of categories of characters or the thresholds of radical frequency in the training set.}
    \label{table:zeroshot_recognition}
\end{table*}

\subsection{Experiments on Unseen Character Recognition}
Zero-shot learning is a term that is commonly used in computer vision to refer to the ability to generalize to unseen categories in image classification~\cite{lampert2009}. 
However, in the context of Chinese characters, the term is used in a broader sense to study the generalization to characters with unseen radicals. 
We compare \ours with 4 state-of-the-art zero-shot recognition methods, including four radical-based approaches (DenseRAN~\cite{wang2018}, HDE~\cite{cao2020}, CUE~\cite{luo2023}, ), a stroke-based approach (SDN~\cite{chen2021}, CCR-CLIP~\cite{yu2023}) and a compound approach (STAR~\cite{zeng2023}).

In particular, zero-shot experiments are conducted on two datasets in two settings, as shown in Table.~\ref{table:zeroshot_recognition}, including character zero-shot and radical zero-shot. 
During testing, all categories that appear in the corresponding experiments are used as candidates.

\paragraph{Character Zero-shot Settings}
The selection of categories in the training set and testing set has a significant impact on Chinese character zero-shot recognition experiments~\cite{cao2020}. 
To ensure a fair comparison, we use the same sample division strategy as~\cite{yu2023,chen2021}. 
For handwritten characters, we first sort the 3755 level-1 commonly used Chinese characters according to the order of GB2312. 
Then, we choose samples from CASIA-HWDB with labels in the first $m$ characters as the training set, where $m\in[500,\ 1000,\ 1500,\ 2000,\ 2755]$. 
Finally, samples from ICDAR2013 with labels in the last $1000$ characters are chosen as the test set.
For scene characters, we first sort all 3650 characters that appear in CTW according to the order of GB18030-2005. 
Then, we collect the samples in the first $500$ characters as the test set and the following m characters as the training set, where $m\in[500,\ 1000,\ 1500,\ 2000,\ 3150]$.

The experimental results reported in the left part of Table.~\ref{table:zeroshot_recognition}.
In most cases, CLIP-based approaches outperforms other approaches by a significant margin.
This is because the CLIP-based method models both the image and the description sequence, whereas the other methods only model the image and require hard matching to find objects in the dictionary that are similar to the generated sequences.
It is worth noting that although the recognition accuracy has been improved by a large margin through the CLIP model, our method still significantly improves the recognition accuracy by effectively utilizing the tree structure of the radical-based sequences.
These results demonstrate the effectiveness of the proposed method.

We only failed to achieve best performance in the zero-shot experiment with training samples of 500 scene characters, which may be due to insufficient training samples or the need for data augmentation.
The experimental results show that the proposed \ours model outperforms the compared methods in most character zero-shot settings.

\paragraph{Radical Zero-shot Settings}
The division of characters for radical zero-shot is accomplished in two steps. 
First, the frequency of each radical is calculated in the corresponding dataset. 
Then, if a sample's corresponding character contains a radical that appears less than n times, where $n\in[50,\ 40,\ 30,\ 20,\ 10]$, it is moved to the test set; otherwise, it is moved to the training set.

The idea of radical zero-shot is proposed to demonstrate the advantage of stroke-based methods~\cite{chen2021} over radical-based methods~\cite{wang2018, cao2020, yu2023}. 
Despite our text encoder's inability to obtain precise representations of unseen radicals, our radical-based method still outperforms existing methods by a significant margin. 
The possible reason for this is that our tree-based representation effectively integrates the radical-based formation characteristics of Chinese characters.

\begin{table}[b!]
    \small
    \centering
    \begin{tabular}{llll}
    \hline
    Method                              & ICDAR2013         & CTW               & AIT (ms)          \\  \hline
    ResNet~\cite{he2016}                & 96.83\%           & 79.46\%           & 12                \\
    Human~\cite{yin2013}                & 96.13\%           & $-$               & $-$               \\
    DenseRAN~\cite{wang2018}            & 96.66\%           & 85.56\%           & 1666              \\
    SDN~\cite{chen2021}                 & 96.28\%           & 85.29\%           & 567               \\
    RTN~\cite{xue2023}                  & 96.88\%           & 88.64\%           & $-$               \\
    STAR~\cite{zeng2023}                & 97.11\%           & 85.43\%           & $-$               \\
    HDE~\cite{cao2020}                  & 97.14\%           & \textbf{89.25}\%  & 29                \\
    CCR-CLIP~\cite{yu2023}              & 97.18\%           & 85.78\%           & 14                \\ 
    RZCR~\cite{diao2023}                & 97.42\%           & 88.74\%           & $-$               \\  \hline
    \ours                               & \textbf{97.87}\%  & 87.82\%           & 0.31              \\  \hline
    \end{tabular}
    \caption{Comparison of performance and average inference time(AIT) on ICDAR-2013 competition set and CTW set with other state-of-theart methods.}
    \label{table:character_recognition}
\end{table}

\begin{table*}[t]
        \renewcommand{\arraystretch}{1.2}
        \small
        \centering
        \begin{tabular}{ccccclllll}
            \hline
            Approach            & Attention Bias & PE      & Special Node & Tree Mask  & $m$=500          & $m$=1000         & $m$=1500         & $m$=2000        & $m$=2755         \\ \hline
            \multirow{2}{*}{Sequential} & $-$            & Learned & [CLS]        & $-$        & 15.6\%           & 35.75\%          & 51.19\%          & 58.85\%         & 69.57\%          \\
                                      & $-$            & Azimuth & [CLS]        & $-$        & 14.93\%          & 33.7\%           & 48.48\%          & 59.05\%         & 68.93\%          \\ \hline
            \multirow{4}{*}{Tree-based}     & SubTree        & $-$     & $-$          & $-$        & 21.18\%          & 43.84\%          & 59.91\%          & 66.12\%         & 75.73\%          \\
                                      & SubTree        & $-$     & $-$          & \checkmark & 22.82\%          & 47.34\%          & 60.65\%          & 67.15\%         & 77.38\%          \\
                                      & SubTree        & Azimuth & [VNODE]      & \checkmark & 24.93\%          & 49.27\%          & 62.14\%          & \textbf{71.6\%} & \textbf{79.78\%} \\
                                      & SubTree        & Azimuth & $-$          & \checkmark & \textbf{26.17\%} & \textbf{50.68\%} & \textbf{62.25\%} & 70.39\%         & 78.96\%          \\ \hline
        \end{tabular}
        \caption{Ablation study of our proposed methods on character zero-shot recognition, $m$ indicates the total number of categories of characters are served as training samples. PE indicates positional encoding used in Transformers.}
        \label{table:ftree_ablation}
\end{table*}
\subsection{Experiments on Seen Character Recognition}
We also train the proposed \ours using all training samples in seen character recognition, where all characters in the test dataset are covered by the training dataset.
For handwritten characters, we use CASIA-HWDB as the training set and ICDAR2013 as the test set.
As shown in Table~\ref{table:character_recognition}, the experimental results indicate that the proposed method achieves the best performance on the handwritten character dataset, and our lightweight model significantly outperforms existing models in computational efficiency. 
However, our model does not lead in the scene character dataset, suggesting the image encoder needs further refinement for processing Chinese characters in complex scenes.
It is worth mentioning that we achieved this performance with a lightweight model. 
In particular, compared to CCR-CLIP, which uses a larger transformer-based text encoder with 12 layers and a hidden size set to 1024, our tree encoder has only 8 layers and a hidden size of 512. 
Moreover, our image encoder is also much smaller relative to their ResNet counterparts.

\subsection{Ablation Studies}
To assess the performance improvements offered by our proposed methods, we conducted ablation studies focused on zero-shot character recognition experiments. 
Both the sequential and tree-based approaches provide unambiguous representations of Chinese characters and show similar effectiveness in recognizing seen characters. However, they exhibit significant differences in their ability to recognize unseen characters.
To facilitate a fair comparison across different configurations while managing memory constraints, we standardized the batch size at 1024 for all experiments, ensuring uniformity in parameter setting.

\paragraph{SubTree Encoding}
Our approach relies on the radical-based formation tree as an ideal data structure for representing Chinese characters.
The difference between processing sequences or formation trees in our approach is whether SubTree Encoding is used.
Our tree encoder with SubTree Encoding outperforms the counterpart treating input as sequences, demonstrating the advantage of our formation tree over sequences for representing Chinese characters as shown in Table~\ref{table:ftree_ablation}.
The results that using sequential approaches obtain relatively poor accuracy, which indicates that the formation tree just effectively combines the Chinese character domain knowledge into the model to further improve the performance.

\paragraph{Azimuth Encoding}
When treating input as formation trees, our tree encoder with Azimuth Encoding yields a decent performance boost compared to ones without Azimuth Encoding as shown in Table~\ref{table:ftree_ablation}.
This indicates that Azimuth Encoding is indispensable to our tree encoder for modeling our formation trees.
However, a similar performance boost does not appear when treating input as sequences, indicating that the Azimuth Encoding alone cannot provide complete structural features for sequential approach.

\paragraph{Tree Mask}
We masked out all the unseen leaf nodes(radicals) in the formation tree in the inference stage.
As mentioned in above section, we tried to improve radical zero-shot recognition by replacing leaf nodes with [unk] nodes or masking leaf nodes randomly during training did not yield the expected improvements.
However, masking unseen nodes during inference effectively improves recognition accuracy as shown in Table.~\ref{table:ftree_ablation}.
This demonstrates that excluding the effect of unseen nodes on tree representations during inference can significantly aids in leveraging known radical features to match image-tree pairs accurately. 

\paragraph{Special Node}
We compared our formation tree representation learning with the root node with the commonly used Special Node.
When using the sequential approaches, we use the [CLS] token as used in BERT~\cite{devlin2019} and we employing tree-based approaches, we utilize the [VNode] node same as Graphormer~\cite{ying2021}.
As shown in Table.~\ref{table:ftree_ablation}, the performance gap between the adding special node or not is minor for tree-based approaches, but using the root node will not add additional compute costs.
This suggests that it is better to use the root node to learn the global representation of the formation tree.

\begin{figure*}[!t]
\centering
\subfloat[$m$=500]{\includegraphics[width=0.32\textwidth]{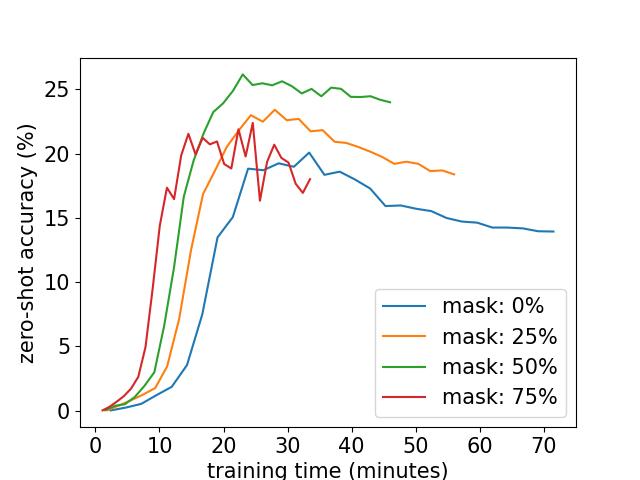}%
\label{fig_zs_500}}
\hfil
\subfloat[$m$=1000]{\includegraphics[width=0.32\textwidth]{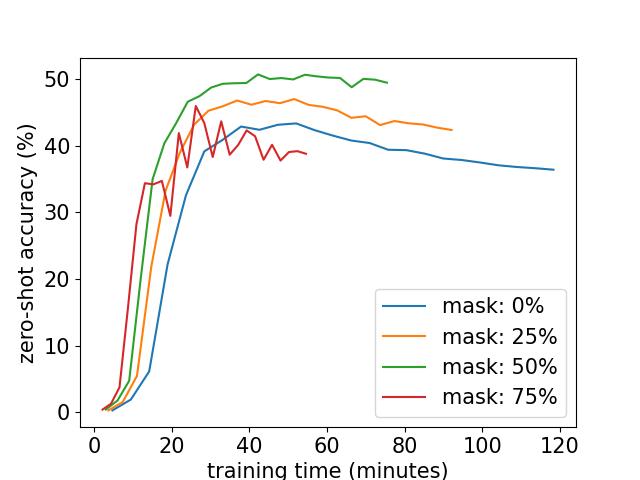}%
\label{fig_zs_1000}}
\hfil
\subfloat[$m$=1500]{\includegraphics[width=0.32\textwidth]{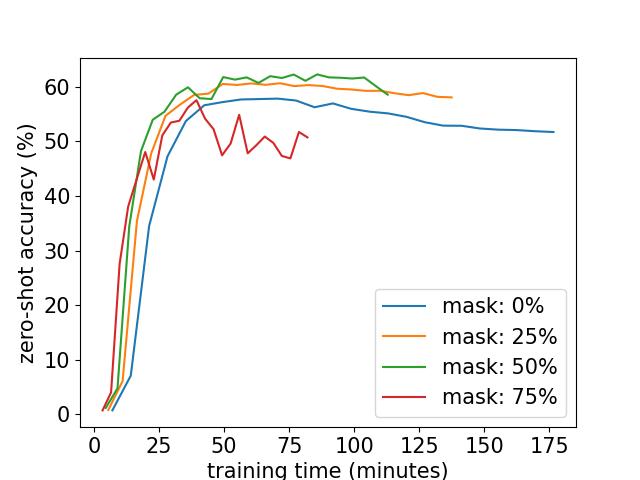}%
\label{fig_zs_1500}}
\hfil
\subfloat[$m$=2000]{\includegraphics[width=0.32\textwidth]{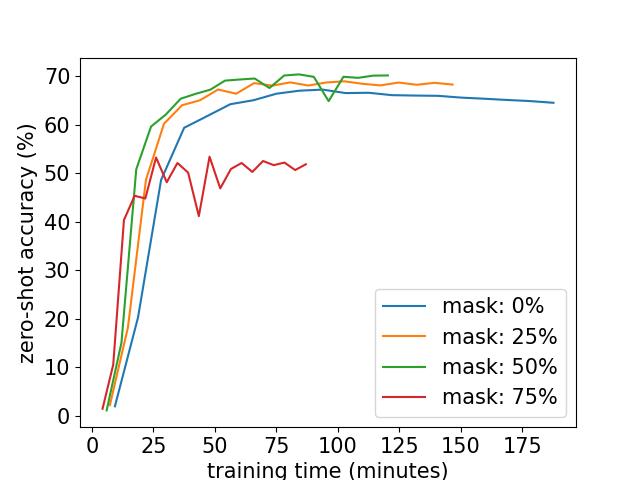}%
\label{fig_zs_2000}}
\hfil
\subfloat[$m$=2755]{\includegraphics[width=0.32\textwidth]{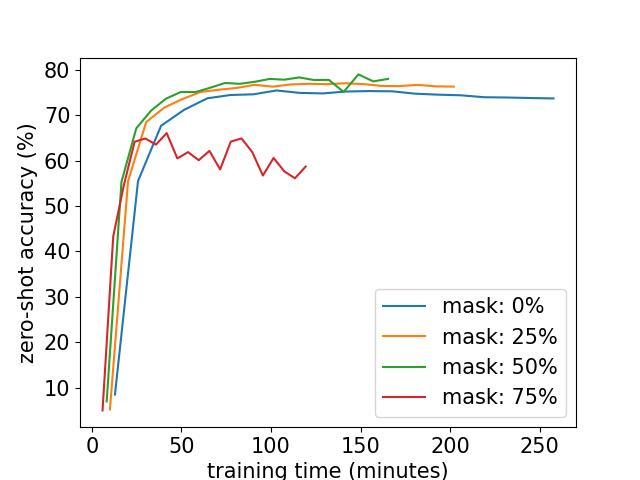}%
\label{fig_zs_2755}}
\caption{Accuracy vs. training time trade-off by image mask ration in detail.
Four kinds of mask ratios are tested in five different cases of character zero-shot recognition tasks. 
Different mask ratios are marked with different colors.
$m$ indicates the total number of categories of characters are served as training samples.
In each group of experiments, all parameters remain constant, except for the mask ratio.}
\label{fig:mask_ratio}
\end{figure*}

\paragraph{Image Mask Ratio} 
Our image encoder incorporates random masking to enhance computational efficiency and effectiveness.
As depicted in Fig.~\ref{fig:mask_ratio}, our method achieves the best performance and less time consumption with a mask ratio of 50\%. 
However, the results of 75\% masking fail to achieve ideal performance as expected in MAE~\cite{he2022}, although they obtain the largest computational efficiency gain.
The possible reasons for this situation may be that Chinese character images are more detailed than natural images, a large mask ratio will make them difficult to identify. 
The results of 25\% masking not only do not get the best performance, but also have a limited improvement in improving computational efficiency. 
In each group of experiments, all parameters remain constant, except for the mask ratio.

\section{Conclusion}
This paper presents a novel approach, Formation Tree-CLIP (\ours), that leverages the inherent tree structure of radical-based sequences to enhance Chinese character representation. Inspired by CLIP, \ours learns transferable character representations. The core components of \ours include: 
\textbf{Formation Tree Representation}: We introduce a novel formation tree to represent the radical-based structure of Chinese characters. This representation captures the hierarchical relationships between radicals.
\textbf{Dedicated Tree Encoder}: To effectively process these non-Euclidean formation trees, \ours incorporates a dedicated tree encoder with two simple yet efficient encoding methods.
\textbf{Masking for Efficiency}:To enable efficient and effective training of large models, \ours utilizes masking techniques applied to both character images and tree nodes.
Extensive evaluations demonstrate that \ours achieves state-of-the-art performance across a wide range of experimental setups, encompassing both seen and unseen character recognition tasks on widely used benchmark datasets. Notably, this superior performance is achieved with a lightweight model that is significantly faster (over ten times) compared to the most efficient existing approach. 
Future work will explore the applicability of the learned character representations to other Chinese character-related tasks.

{
    \small
    \bibliographystyle{ieeenat_fullname}
    \bibliography{ftclip}
}


\end{document}